\documentclass{article}

\usepackage[final]{corl_2019} 

\usepackage{microtype}
\usepackage{graphicx}
\usepackage{subfigure}
\usepackage{booktabs} 
\usepackage{multicol}
\usepackage{wrapfig}

\usepackage{amsmath}
\usepackage{amsfonts}
\usepackage{mathtools}

\usepackage{hyperref}

\usepackage[dvipsnames]{xcolor}
\usepackage[utf8]{inputenc}
\DeclareUnicodeCharacter{2212}{-}

\usepackage{pgfplots}
\usepackage{tikz} 

\pgfplotsset{compat=newest}
\usetikzlibrary{decorations.pathmorphing} 
\usetikzlibrary{fit} 
\usetikzlibrary{backgrounds} 
\usetikzlibrary{pgfplots.groupplots}
\usetikzlibrary{positioning}
\usetikzlibrary{shapes.misc}

\tikzset{every axis/.append style = {
    major tick length=2.5,
    every tick/.style={
            black,
    }
    }
}

\tikzset{font={\fontsize{8pt}{9.6pt}\selectfont}}

\tikzset{cross/.style={cross out, draw=black, minimum size=2*(#1-\pgflinewidth), inner sep=0pt, outer sep=0pt, very thick}, cross/.default={1pt}}

\usepackage{algorithm}
\usepackage{algorithmic}

\newcommand{\cvec}[1]{\boldsymbol{#1}}
\newcommand{\bindent}[1]{%
  \begingroup
  \setlength{\itemindent}{#1}
}
\newcommand{\eindent}{\endgroup}

\DeclareMathOperator*{\argmax}{arg\,max}

\newcommand{\norm}[1]{\left\lVert#1\right\rVert}

\makeatletter
\newcommand\mathlabel[1]{&\refstepcounter{equation}(\theequation)\ltx@label{#1}&}
\makeatother

\title{Self-Paced Contextual Reinforcement Learning}

\author{
  Pascal Klink\\
  Intelligent Autonomous Systems\\
  Technische Universität Darmstadt\\
  Germany\\
  \texttt{klink@ias.tu-darmstadt.de} \\
  \And
  Hany Abdulsamad\\
  Intelligent Autonomous Systems\\
  Technische Universität Darmstadt\\
  Germany\\
  \texttt{abdulsamad@ias.tu-darmstadt.de} \\
  \And
  Boris Belousov\\
  Intelligent Autonomous Systems\\
  Technische Universität Darmstadt\\
  Germany\\
  \texttt{belousov@ias.tu-darmstadt.de} \\
  \And
  Jan Peters\\
  Intelligent Autonomous Systems\\
  Technische Universität Darmstadt\\
  Germany\\
  \texttt{peters@ias.tu-darmstadt.de} \\
}

\begin{document}
\maketitle


\begin{abstract}
	Generalization and adaptation of learned skills to novel situations
	is a core requirement for intelligent autonomous robots.
	Although contextual reinforcement learning provides a principled framework
	for learning and generalization of behaviors across related tasks,
	it generally relies on uninformed sampling of environments
	from an unknown, uncontrolled context distribution,
	thus missing the benefits of structured, sequential learning.
	We introduce a novel relative entropy reinforcement learning algorithm
	that gives the agent the freedom to control the intermediate task distribution,
	allowing for its gradual progression towards the target context distribution.
	Empirical evaluation shows that the proposed curriculum learning scheme
	drastically improves sample efficiency and enables learning
	in scenarios with both broad and sharp target context distributions
	in which classical approaches perform sub-optimally.
\end{abstract}



\section{Introduction}
\label{sec:intro}
Reinforcement learning (RL) techniques underpinned the recent successes
in decision making \citep{mnih-atari, silver-alpha-go-zero}
and robot learning \citep{kober2009policy, levine2016end}.
While most impressive results were achieved in the classical single-task RL setting
with a static environment and a fixed reward function,
contextual reinforcement learning holds the promise of driving the next wave of breakthroughs
by leveraging similarities between environments and tasks
\citep{schaul-uvfas, andrychowicz-her, pong-temporal-difference-models}.
Such a high-level abstraction of shared task structure permits simultaneous multi-task policy optimization
and generalization to novel scenarios through interpolation and extrapolation
\citep{neumann-creps, ammar-multi-task-policy-gradients, kupcsik-gp-reps, parisi-creps-tetherball}.

A crucial limitation of the contextual RL framework is the assumption that the task distribution
is not known beforehand and not controlled by the agent.
Despite being effective in the contextual bandit setting \citep{lattimore2018bandit},
where the context is chosen by an underlying unknown (adversarial) stochastic process,
this assumption is rather limiting in robotics from a practical and an algorithmic point of view.
On one hand, the target context distribution is usually known to the designer of the system,
and  contextual learning is purposefully applied to generalize a skill over a set of target contexts,
be it the whole context space or a specific part of it \citep{deisenroth-rl-survey}.
On the other hand, not exploiting the freedom in adjusting the context distribution to the skill level of the agent
can only deteriorate the performance of the learning algorithm.
Indeed, advances in gradual and sequential learning, exemplified by shaping \citep{krueger2009flexible}
and curriculum learning \citep{bengio-curriculum-learning},
which are loosely based on the insights from behavioral psychology \citep{skinner1990behavior}
and continuation methods in optimization \citep{allgower-continuation-methods},
demonstrate significant improvements in the speed of learning and resilience to local optima.

Defining an appropriate curriculum is crucial for the success of gradual learning approaches.
For contextual RL, this means repeatedly choosing a batch of tasks
that result in the greatest improvement in the direction of the target contexts.
Unfortunately, finding an optimal curriculum is an intractable sequential decision making problem.
In this paper, we instead construct a curriculum as a series of intermediate context distributions
that we gradually optimize based on a trade-off between local reward maximization
and the expected progress towards the target context distribution.
This allows an agent to pace its learning process autonomously
based solely on the interactions with the environment.

Our formulation of the one-step optimal curriculum selection problem
is instantiated in the framework of contextual relative entropy policy search
(C-REPS) \citep{kupcsik-gp-reps, parisi-creps-tetherball},
offering a principled way of imposing constraints through the Lagrangian dual formulation. Nonetheless the proposed approach can be adapted to enhance other contextual RL solvers too.
In the following sections, we introduce our algorithm and empirically investigate its performance and sample efficiency on three robotics tasks, among them an especially challenging sparse Ball-in-a-Cup task on a Barrett WAM, demonstrating that a self-paced curriculum facilitates learning by drastically reducing sample complexity in comparison to the baseline methods.
Finally, we highlight and discuss related RL and curriculum learning approaches,
pointing out promising directions for future work.

\section{Contextual Reinforcement Learning}
\label{sec:creps}
We formulate classical contextual RL as a stochastic search problem in a continuous space of contexts $\cvec{c} \in C$, that define the configuration of an environment, and a space of policy parameters $\cvec{\theta} \in \Theta$, which dictate the agent's behavior. A reward function $R(\cvec{\theta}, \cvec{c})$ maps every point of the product space to a real number, $R: C \times \Theta \rightarrow \mathbb{R}$. The context $\cvec{c}$ is assumed to be drawn from an unknown distribution $\mu(\cvec{c})$, while the policy parameters are sampled from a conditional search distribution $\pi({\cvec\theta}|\cvec{c})$. Our definition motivates the stochastic search objective
$ J(\pi) = \int_{C} \mu(\cvec{c}) \int_{\Theta} \pi(\cvec{\theta} \vert \cvec{c}) R(\cvec{\theta}, \cvec{c}) \text{d}\cvec{\theta} \text{d}\cvec{c}.$

Contextual relative entropy policy search (C-REPS) frames the search problem as an iterative entropy-regularized optimization under the distribution $p(\cvec{\theta}, \cvec{c}) = \mu(\cvec{c}) \pi(\cvec{\theta} \vert \cvec{c})$,
\begin{align*}
	\argmax_{p({\theta}, \cvec{c})} \quad & \int_{{C}, {\Theta}} R(\cvec{\theta}, \cvec{c}) p(\cvec{\theta}, \cvec{c}) \text{d}\cvec{c} \text{d}\cvec{\theta},
	\qquad \text{s.t.} \quad  \text{D}_{\text{KL}}\left(p(\cvec{\theta}, \cvec{c}) || q(\cvec{\theta}, \cvec{c})\right) \leq \epsilon,                         \\
	                                      & \int_{C, \Theta} p(\cvec{\theta}, \cvec{c}) \text{d}\cvec{c} \text{d}\cvec{\theta} = 1,
	\qquad  \qquad \quad  \int_{\Theta} p(\cvec{\theta}, \cvec{c}) \text{d}\cvec{\theta} = \mu(\cvec{c}), \quad \forall \cvec{c} \in {C},
\end{align*}
where $\text{D}_{\text{KL}}\left(p(.) || q(.)\right)$ is the Kullback-Leibler (KL) divergence  between the distribution $p$ being optimized and the previously found joint distribution $q$. This constraint controls the exploration-exploitation trade-off via the hyperparameter $\epsilon$ and limits information loss between iterations. The remaining constraints are necessary to ensure distribution normalization and marginalization properties. This constrained optimization problem is tackled by formulating the Lagrangian function and solving the primal problem, which yields the following optimality condition for $p(\cvec{\theta}, \cvec{c})$
\begin{equation}
	p^{*}(\cvec{\theta}, \cvec{c}) \propto q(\cvec{\theta}, \cvec{c})\exp\left(\frac{A(\cvec{\theta}, \cvec{c})}{\eta}\right),
	\label{eq:creps_opt}
\end{equation}
which can be interpreted as soft-max re-weighting of the joint distribution $q$ by the advantage function $A\left(\cvec{\theta}, \cvec{c}\right) = R(\cvec{\theta}, \cvec{c}) - V(\cvec{c})$. The parameters $\eta$ and $V(\cvec{c})$ are the Lagrangian variables corresponding to the KL and marginalization constraints respectively. Those parameters can be optimized by minimizing the corresponding dual objective, which is derived by plugging the optimal distribution $p^{*}(\cvec{\theta}, \cvec{c})$ back into the primal problem,
\begin{equation*}
	\mathcal{G}(\eta, V) = \eta \epsilon + \eta \log\left(\mathbb{E}_q \left[\exp\left(\frac{A(\cvec{\theta}, \cvec{c})}{\eta} \right) \right]\right).
\end{equation*}
Given that neither the sampling distribution $q(\cvec{\theta}, \cvec{c})$ nor the advantage $A\left(\cvec{\theta}, \cvec{c}\right)$ are known analytically, the dual objective can only be approximated under samples from $q(\cvec{\theta}, \cvec{c})$. By assuming a certain function class for the optimal policy $\pi^{*}(\cvec{\theta}|\cvec{c})$, for example a Gaussian process or an RBF network, the optimality condition in Equation~\ref{eq:creps_opt} can be satisfied by performing a maximum a posteriori fit under samples drawn from $q(\cvec{\theta}|\cvec{c})$ and $\mu(\cvec{c})$.

\clearpage

		\begin{algorithm}[H]
	\caption{Self-Paced Reinforcement Learning}
	\label{alg:sprl}
	\begin{algorithmic}
		\STATE{\bfseries Input:} Relative entropy bound $\epsilon$,
		offset $K_{\alpha}$ (after which $\alpha$ becomes non-zero),
		KL-penalty fraction  $\zeta$,
		initial policy $q_{0}(\cvec{\theta} \vert \cvec{c})$,
		initial sampling distribution $q_{0}(\cvec{c})$,
		number of iterations $K$.
		
		\FOR{$k=1$ {\bfseries to} $K$}
		\STATE {\bfseries Collect Data:}
		\bindent{1em}
		\STATE Sample contexts: $\cvec{c}_{i} \sim q_{k-1}(\cvec{c}),\ i = 1, \ldots, M$
		\STATE Sample and execute actions: $\cvec{\theta}_{i} \sim q_{k-1}(\cvec{\theta} \vert \cvec{c}_{i})$
		\STATE Observe reward: $R_i = R\left(\cvec{\theta}_i, \cvec{c_i}\right)$
		\eindent{}
		
		\STATE {\bfseries Update Policy and Context Distributions:}
		\bindent{1em}
		
		\STATE Update schedule: $\alpha_k = 0,\ \text{if } k \leq K_{\alpha},\ \text{else } \frac{\zeta \sum_{i =1}^M R_i}{M D_{\mathrm{KL}}(q_{k - 1} \| \mu)}$
		\STATE Optimize dual function:  $\left[\eta_{\mu}^*, \eta_{p}^*, V^*\right] \leftarrow \argmax \mathcal{G}(\eta_{\mu}, \eta_{p}, V)$
		
		\STATE Calculate sample weights: $\left[w^{\pi}_i, w^{\tilde{\mu}}_i\right] \leftarrow \left[ \exp\left(\frac{A(\cvec{\theta}_i, \cvec{c}_i)}{\eta_{p}^*}\right), \exp\left(\frac{\beta(\cvec{c}_i)}{\alpha_k + \eta_{\mu}^*}\right)\right]$
		
		\STATE Infer policy and context distributions: $\left[\pi^{*}_{k}, \tilde{\mu}^{*}_{k}\right] \leftarrow D = \left\{(w^{\pi}_i, w^{\tilde{\mu}}_i, \cvec{\theta}_{i}, \cvec{c}_i) \vert i \in \left[1, M\right] \right\}$

		\STATE Assign policy and context distributions: $q_{k}(\cvec{\theta}| \cvec{c}) \leftarrow \pi^{*}_{k}(\cvec{\theta}| \cvec{c}), ~q_{k}(\cvec{c}) \leftarrow \tilde{\mu}^{*}_{k}(\cvec{c})$
		\eindent{}
		\ENDFOR
	\end{algorithmic}
\end{algorithm}

\section{Self-Paced Reinforcement Learning}
\label{sec:sprl}

Following our line of argument in Section~\ref{sec:intro},
we want to transform the contextual stochastic search problem
presented in Section~\ref{sec:creps} from a passive optimization under an unknown distribution $\mu(\cvec{c})$ into one that allows the learning agent to actively optimize an intermediate context distribution $\tilde{\mu}(\cvec{c})$ and initially focus on ``easy" tasks and gradually progress towards the target distribution $\mu(\cvec{c})$ by bootstrapping the optimal policy from the previous iterations.

This progression is to be understood as a trade-off between maximizing the objective $J=\mathbb{E}_{\pi, \tilde{\mu}}\left[R(\cvec{\theta}, \cvec{c})\right]$ by exploiting local information and minimizing the distance to the context distribution of interest ${\mu}(\cvec{c})$. A connection to numerical continuation methods \citep{allgower-continuation-methods} can be drawn by viewing the intermediate distribution $\tilde{\mu}(\cvec{c})$ as a parameter of the main objective $J(\pi, \tilde{\mu})$. Under this view, the distribution $\tilde{\mu}(\cvec{c})$ gradually morphs the function $J(\pi)$ to improve convergence when a good initial guess of $\pi$ is not available.

In the following, we present our self-paced contextual RL (SPRL) algorithm, which incorporates such intermediate distributions into the framework of C-REPS and allows an agent to optimize its own learning process, under the assumption of a known target context distribution.

We formulate a new constrained optimization problem that takes into consideration a ``local" objective of maximizing the expected reward and a ``global" objective seeking to move $\tilde{\mu}(\cvec{c})$ towards ${\mu}(\cvec{c})$ by minimizing $\text{D}_{\text{KL}}\left(\tilde{\mu} || \mu \right)$,
\begin{align}
	\argmax_{p, \tilde{\mu}} \quad & \int_{C, \Theta} R(\cvec{\theta}, \cvec{c}) p(\cvec{\theta}, \cvec{c}) \text{d}\cvec{c} \text{d}\cvec{\theta} - \alpha \text{D}_{\text{KL}}\left(\tilde{\mu} || \mu \right) \\
	\text{s.t.} \quad              & \text{D}_{\text{KL}}\left(p(\cvec{\theta}, \cvec{c}) || q(\cvec{\theta}, \cvec{c}) \right) \leq \epsilon \label{eq:kl_pq},                                                  \\
	                               & \int_{C, \Theta} p(\cvec{\theta}, \cvec{c}) \text{d}\cvec{c} \text{d}\cvec{\theta} = 1, \label{eq:p_norm}                                                                   \\
	                               & \int_{\Theta} p(\cvec{\theta}, \cvec{c}) \text{d}\cvec{\theta} = \tilde{\mu}(\cvec{c}) , ~\forall \cvec{c} \in C \label{eq:p_mu},
\end{align}
where the joint distributions $p(.)$ and $q(.)$ are defined as $p(\cvec{\theta}, \cvec{c}) = \pi(\cvec{\theta} \vert \cvec{c}) \tilde{\mu}(\cvec{c})$ and $q(\cvec{\theta}, \cvec{c}) = q(\cvec{\theta} \vert \cvec{c}) q(\cvec{c})$, respectively.
It follows that in every iteration the learning agent is able to optimize both its policy $\pi(\cvec{\theta} \vert \cvec{c})$ and sampling distribution $\tilde{\mu}(\cvec{c})$, starting from the uninformative Gaussian priors $q(\cvec{c})$ and $q(\cvec{\theta}|\cvec{c})$, while following a schedule of $\alpha$.

\subsection{Optimality Conditions and Dual Optimization}

\begin{figure*}[t]
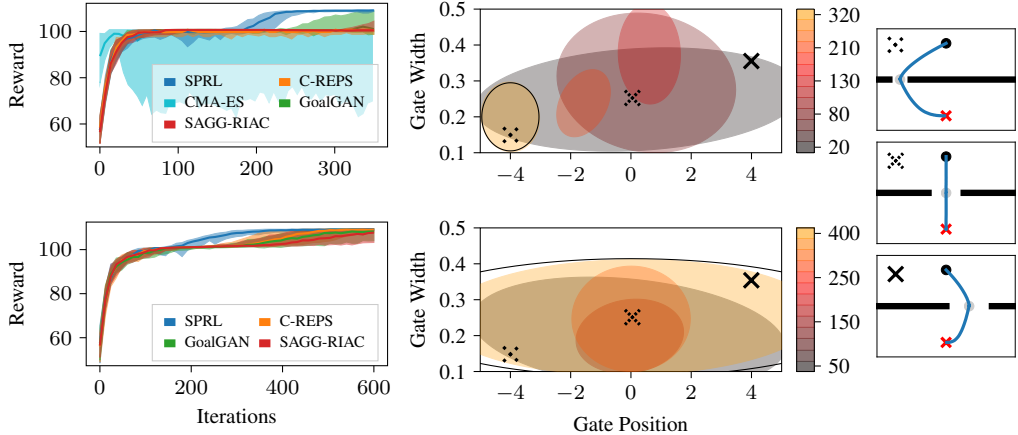

	\centering
	\begin{tikzpicture}
		\node (O) {\input{./gate-precision+gate-global-rewards.tex}};
		\node (O0) [right = -5pt of O] {\input{./gate-precision+gate-global-distribution-evolution.tex}};
		\node [above right = -60pt and -5pt of O0] (O1) {\scalebox{0.12}{\input{./gate-pos-1.pgf}}};
		\node [below = -5.8pt of O1] (O2) {\scalebox{0.12}{\input{./gate-pos-3.pgf}}};
		\node [below = -5.8pt of O2] (O3) {\scalebox{0.12}{\input{./gate-pos-2.pgf}}};

		\node[cross=4pt, dotted, above left = -14pt and -15pt of O1] (C1) {};
		\node[cross=4pt, dotted, below left = -38pt and -47pt of O0] (C1) {};
		\node[cross=4pt, dotted, below left = -121pt and -47pt of O0] (C1) {};

		\node[cross=4pt, dash pattern={on 1pt off 0.75pt}, above left = -15pt and -15pt of O2] (C1) {};
		\node[cross=4pt, dash pattern={on 1pt off 0.75pt}, below left = -52pt and -93pt of O0] (C1) {};
		\node[cross=4pt, dash pattern={on 1pt off 0.75pt}, below left = -135pt and -93pt of O0] (C1) {};

		\node[cross=4pt, black, above left = -15pt and -15pt of O3] (C1) {};
		\node[cross=4pt, black, below left = -66pt and -138pt of O0] (C1) {};
		\node[cross=4pt, black, below left = -149pt and -138pt of O0] (C1) {};
	\end{tikzpicture}
	\vspace{-5pt}
	\caption[Gate Experiments]
	{The left plots show the reward achieved by the algorithms on the ``precision'' (top row) and ``global'' setting (bottom row) on the target context distributions in the gate environment. Thick lines represent the $50\%$-quantiles and shaded areas the intervals from $10\%$- to $90\%$-quantile of $40$ algorithm executions. One iteration consists of $100$ policy rollouts. The right plot shows the evolution of the sampling distribution $\tilde{\mu}\left(\cvec{c}\right)$ (colored areas) of one run together with the target distribution $\mu\left(\cvec{c}\right)$ (black line). The small images on the right visualize the task for different gate positions and widths. The crosses mark the corresponding positions in the context space.}
	\label{fig:gate-experiments}
	\vspace{-5pt}
\end{figure*}

By constructing the Lagrangian function of the above described optimization problem, we can derive the optimality conditions for $p^{*}(\cvec{\theta}, \cvec{c})$ and $\tilde{\mu}^{*}(\cvec{c})$. However, during experiments we observed that the resulting expressions and the dual objective are numerically unstable for small values of~$\alpha$. A more detailed treatment of those problems can be found in the supplementary material. We can avoid such numerical issues by over-defining the original problem and adding the following conditions, which are implicitly satisfied when the constraints in Equations~(\ref{eq:kl_pq}-\ref{eq:p_mu}) hold,
\begin{align*}
	 & \int_{C} \tilde{\mu}\left(\cvec{c}\right) \text{d}\cvec{c} = 1 \mathlabel{eq:mu_norm}, &  & \text{D}_{\text{KL}}\left(\tilde{\mu}(\cvec{c}) || q(\cvec{c})\right) \leq \epsilon. \mathlabel{eq:kl_mu} &
\end{align*}
Solving the new augmented primal problem results in an optimality condition $p^{*}(\cvec{\theta}, \cvec{c})$ which is equivalent to that in Equation~\eqref{eq:creps_opt}, while the optimal point $\tilde{\mu}^{*}(c)$ is given by
\begin{align}
	\tilde{\mu}^*(\cvec{c}) & \propto q(\cvec{c})\exp\left(\frac{\alpha \log\left(\frac{\displaystyle \mu(\cvec{c})}{\displaystyle q(\cvec{c})}\right)+V(\cvec{c})}{\alpha+\eta_{\mu}}\right) = q(\cvec{c})\exp\left(\frac{\beta(\cvec{c})}{\alpha+\eta_{\mu}}\right), \label{eq:sprl_opt}
\end{align}
where $\eta_{p}$, $\eta_{\mu}$ and $V(\cvec{c})$  are the Lagrangian multipliers corresponding to Equations~(\ref{eq:kl_pq}, \ref{eq:kl_mu}, \ref{eq:p_mu}) respectively, and are optimized by minimizing the dual objective
\begin{align}
	\mathcal{G} & = (\eta_{p} + \eta_{\mu})\epsilon + \eta_{p}\log\left(\mathbb{E}_{q}\left[\exp\left(\frac{A(\cvec{\theta}, \cvec{c})}{\eta_{p}}\right)\right]\right) + (\alpha+\eta_{\mu})\log\left(\mathbb{E}_{q}\left[\exp\left(\frac{\beta(\cvec{c})}{\alpha+\eta_{\mu}}\right)\right]\right) \label{eq:sprl_dual}.
\end{align}
Equation~\eqref{eq:sprl_dual} underscores the contributions of Equations~(\ref{eq:mu_norm},\ref{eq:kl_mu}) to numerical robustness, as they introduce the logarithmic function and the temperature $\eta_{\mu}$ to the second expectation, consequently reducing the severity of numerical overflow issues. Algorithm~\ref{alg:sprl} depicts a general sketch of the overall learning procedure.

\subsection{Interpretation of the Dual Variable $V(\cvec{c})$}

As already highlighted in \citep{deisenroth-rl-survey}, $V(\cvec{c})$ can be interpreted as a context-value function, representing a soft-max operator over the expected reward under the policy $q(\cvec{\theta}|\cvec{c})$. For the case $\alpha = 0$, we obtain
\begin{equation*}
	\tilde{\mu}^*(\cvec{c}) \propto q(\cvec{c})\exp\left(\frac{V(\cvec{c})}{\eta_{\mu}}\right),
\end{equation*}
which reveals that $\tilde{\mu}^*(\cvec{c})$ shifts the contextual variable's modes of density from areas in the context space with lower expected reward to areas with higher expected reward according to $V(\cvec{c})$, while constraints \eqref{eq:kl_pq} and \eqref{eq:kl_mu} limit the rate of such shifts, in order to limit the information loss compared to previous iterations. This insight motivates a slowly increasing schedule of $\alpha$-values, that allows the agent to concentrate on easier tasks at first before guiding it towards harder tasks.

\subsection{Practical Aspects}
Our approach requires to choose a class of (Bayesian) function approximators for representing the value function $V(\cvec{c})$ and policy $\pi(\cvec{\theta}|\cvec{c})$. Moreover, the distributions ${\mu}(\cvec{c})$ and $\tilde{\mu}(\cvec{c})$ need to be pinned down by a parametric form to allow the sample-weighted updates.

We define the following data-set needed to perform the updates in Equations~\eqref{eq:creps_opt} and \eqref{eq:sprl_opt}
\begin{align*}
	D & = \left\{\left( w^{\pi}_i, w^{\tilde{\mu}}_i, \cvec{\theta}_{i}, \cvec{c}_i \right) \vert w^{\pi}_{i} = \exp\left(\frac{A(\cvec{c}_{i})}{\eta_{p}}\right),\ w^{\tilde{\mu}}_{i} = \exp\left( \frac{\beta(\cvec{c}_{i})}{\alpha + \eta_{\mu}} \right),\ i \in \left[1, M\right] \right\}.
\end{align*}
Another important aspect is the choice of $\alpha$, which penalizes the divergence between $\tilde{\mu}$ and $\mu$. While experimenting with the algorithm, choosing $\alpha$ such that the KL-Divergence between current- and target context distribution makes up a certain fraction $\zeta$ of the current average reward demonstrated promising performance by automatically adapting to the performance of the algorithm while requiring only one parameter to be tuned. Finally, the hyperparameter $\epsilon$ may require task-dependent tuning.

\section{Empirical Evaluation}

\begin{figure*}[t]
	\centering
	\begin{tikzpicture}
		\node (O) {\input{./reacher-obstacle-default-rewards.tex}};
		\node (O0) [above right = -106pt and -5pt of O] {\input{./reacher-obstacle-default-distribution-evolution.tex}};
		\node [above right = -37pt and 0pt of O0] (O1) {\scalebox{0.065}{\includegraphics{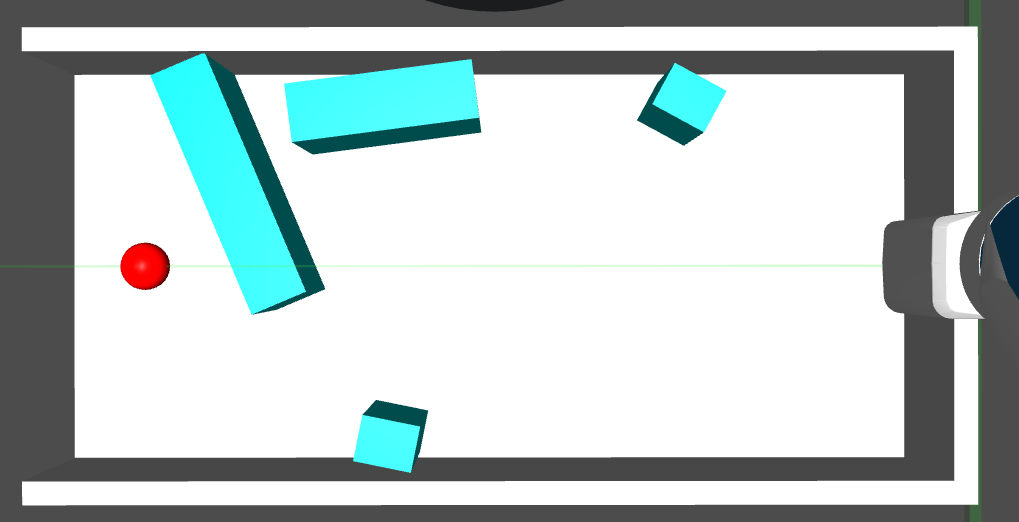}}};
		\node [below = -4pt of O1] (O2) {\scalebox{0.065}{\includegraphics{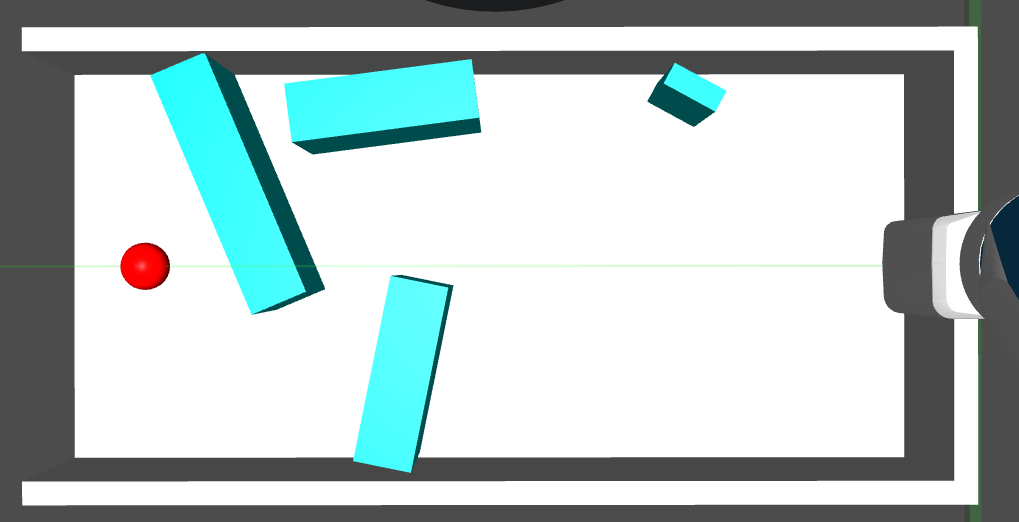}}};
		\node [below = -4pt of O2] (O3) {\scalebox{0.065}{\includegraphics{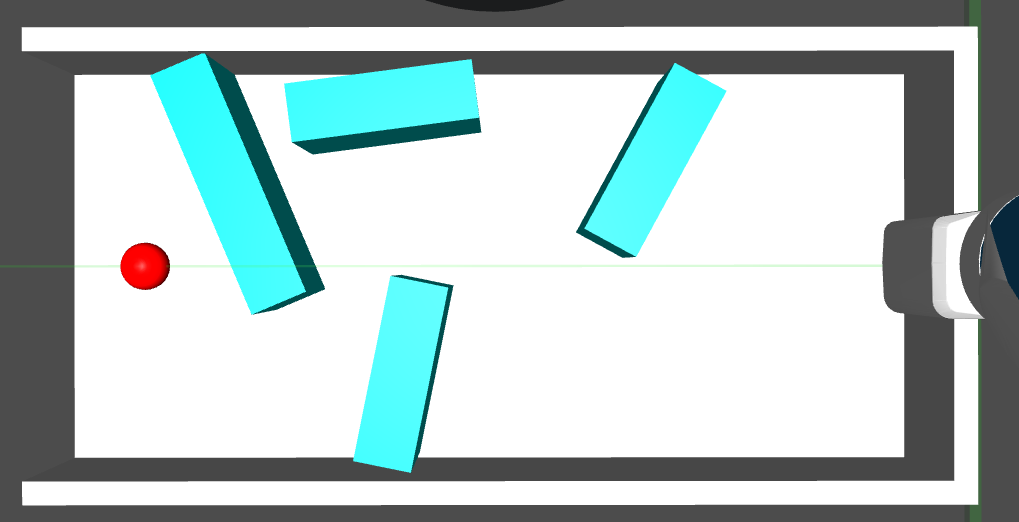}}};

		\node[cross=4pt, dotted, above left = -31pt and -60pt of O1] (C1) {};
		\node[cross=4pt, dotted, below left = -39pt and -41pt of O0] (C1) {};

		\node[cross=4pt, dash pattern={on 1pt off 0.75pt}, above left = -31pt and -60pt of O2] (C1) {};
		\node[cross=4pt, dash pattern={on 1pt off 0.75pt}, above left = -24.5pt and -41pt of O0] (C1) {};

		\node[cross=4pt, black, above left = -31pt and -60pt of O3] (C1) {};
		\node[cross=4pt, black, above left = -24.5pt and -112pt of O0] (C1) {};
	\end{tikzpicture}
	\vspace{-5pt}
	\caption[Avoid Experiments]{
		The left plots show the $50\%$-quantiles (thick lines) and the intervals from $10\%$- to $90\%$-quantile (shaded areas) of the reward achieved by the investigated algorithms on the target context distribution in the Reacher environment. The quantiles are computed from $40$ algorithm executions. One iteration consists of $50$ policy rollouts. Colored areas in the right plot show the sampling distribution $\tilde{\mu}\left(\cvec{c}\right)$ at different iterations of one SPRL run together with the target distribution (black line). The legend on the right shows the iteration that corresponds to a specific color. Small images on the right visualize different contexts with black crosses marking the corresponding positions in context space.}
	\label{fig:avoid-experiments}
	\vspace{-7pt}
\end{figure*}

In this section, we demonstrate the benefit of our algorithm by comparing it to C-REPS, \mbox{CMA-ES}~\citep{hansen-cmaes}, GoalGAN~\citep{florensa-automatic-goal-generation-for-rl} and SAGG-RIAC~\citep{baranes-sagg-riac}. With CMA-ES being a non-contextual algorithm, we only use it in experiments with narrow target distributions, where we then train and evaluate only on the mean of the target context distributions. We will start with a simple point-mass problem, where we evaluate the benefit of our algorithm for broad and narrow target distributions. We then turn towards more challenging tasks, such as a modified version of the reaching task implemented in the OpenAI Gym simulation environment \cite{openai-gym} and a sparse Ball-in-a-Cup task. Given that GoalGAN and SAGG-RIAC are algorithm agnostic curriculum generation approaches, we combine them with C-REPS to make the results as comparable as possible.

In all experiments, we use radial basis function (RBF) features to approximate the value function $V(\cvec{c})$, while the policy $\pi(\cvec{\theta} \vert \cvec{c}) = \mathcal{N}(\cvec{\theta} \vert \cvec{A} \phi(\cvec{c}), \cvec{\Sigma}_{\cvec{\theta}})$ uses linear features $\phi(\cvec{c})$. SPRL and C-REPS always use the same number of RBF features for a given environment. In the case of SPRL, we learn the Gaussian context distribution $\tilde{\mu}(\cvec{c})$ and policy $\pi(\cvec{\theta} \vert \cvec{c})$ jointly from the sample set $D$ while limiting the change in KL-Divergence. For C-REPS, we use the same scheme but only learn the policy from $D$. The approach is outlined in the supplementary material.

In our experiments, SPRL always starts with a wide initial sampling distribution $\tilde{\mu}\left(\cvec{c}\right)$ that, in combination with setting $\alpha = 0$ for the first $K_{\alpha}$ iterations, allows the algorithm to automatically choose the initial tasks on which learning should take place. After the first $K_{\alpha}$ iterations, we then choose $\alpha$ following the scheme outlined in the previous section. Experimental details that cannot be mentioned due to space limitations can be found in the supplementary material. \footnote{Code is publicly available under \url{https://github.com/psclklnk/self-paced-rl}}

\subsection{Gate Environment}

\begin{figure}[t]
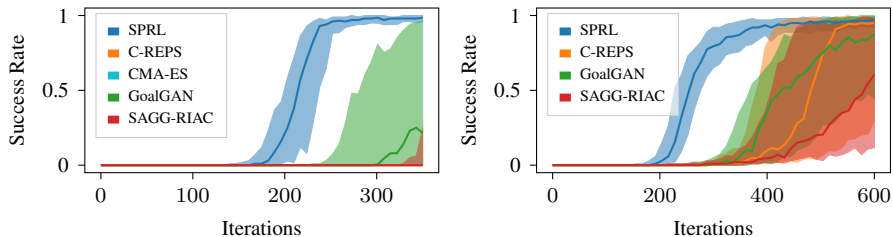

\vspace{-24pt}
	\centering
	\input{./gate-precision+gate-global-successes.tex}
	\input{./gate-precision+gate-global-successes-1.tex}
	\vspace{-5pt}
	\caption[Gate Rewards]{Success rates achieved by the algorithms on the ``precision'' (left) and ``global'' setting (right) of the gate environment. Thick lines represent the $50\%$-quantiles and shaded areas show the intervals from $10\%$- to $90\%$-quantile. The quantiles are computed using $40$ algorithm executions. One iteration consists of $100$ policy rollouts. A policy execution is said to be successful if the final position is sufficiently close to the goal.}
	\label{fig:gate-successes}
	\vspace{-7pt}
\end{figure}

In the first environment, the agent needs to steer a point-mass in two-dimensional space from the starting position $\left[0\ \ 5\right]$ to the goal position at the origin. The dynamics of the point mass are described by a simple linear system subject to a small amount of Gaussian noise. Complexity is introduced by a wall at height $y = 2.5$, which can only be traversed through a gate. The $x$-position and width of the gate together define a task $\cvec{c}$. If the point-mass crashes into the wall, the experiment is stopped and the reward computed based on the current position. The reward function is an exponential of the distance to the goal position with additional L2-Regularization on the generated actions. The point-mass is controlled by two PD-controllers, whose parameters need to be tuned by the agent. The controllers are switched as soon as the point mass reaches the height of the gate, which is why the the desired $y$-position of the controllers are fixed to $2.5$ (the height of the gate) and $0$, while all other parameters are controlled by the policy $\pi$, making $\cvec{\theta}$ a $14$-dimensional vector.

We evaluate two setups in this gate environment, which differ in their target context distribution: In the first one, the agent needs to be able to steer through a very small gate far from the origin (``precision'') and in the second it is required to steer through gates with a variety of positions and widths (``global''). The two target context distributions are shown in Figure~\ref{fig:gate-experiments}.

Figure~\ref{fig:gate-experiments} visualizes the obtained rewards for the investigated algorithms, the evolution of the sampling distribution $\tilde{\mu}\left(\cvec{c}\right)$ as well as sample tasks from the environment. In the ``global'' setting, we can see that SPRL converges significantly faster to the optimum  while in the ``precision'' setting, SPRL avoids a local optimum to which C-REPS and CMA-ES converge and which, as can be seen in Figure~\ref{fig:gate-successes}, does not encode desirable behavior. The visualized sampling distributions in Figure~\ref{fig:gate-experiments} indicate that tasks with wide gates positioned at the origin seem to be easier to solve starting from the initially zero-mean Gaussian policy, as in both settings the algorithm first focuses on these kinds of tasks and then subsequently changes the sampling distributions to match the target distribution. Interestingly, the search distribution of CMA-ES did not always converge in the ``precision'' setting, as can be seen in Figure~\ref{fig:gate-experiments}. This behavior persisted across various hyperparameters and population sizes.

\subsection{Reacher Environment}

For the next evaluation, we modify the three-dimensional Reacher environment of the OpenAI Gym toolkit. In this modified version, the goal is to move the end-effector along the surface of a table towards the goal position while avoiding obstacles that are placed on the table. With the obstacles becoming larger, the robot needs to introduce a more pronounced curve movement in order to reach the goal without collisions. To simplify the visualization of the task distribution, we only allow two of the four obstacles to vary in size. The sizes of those two obstacles make up a task $\cvec{c}$ in this environment. Just as in the first environment, the robot should not crash into the obstacles and hence the movement is stopped if one of the four obstacles is touched. The policy $\pi$ encodes a ProMP ~\citep{paraschos-promps}, from which movements are sampled during training. In this task, $\cvec{\theta}$ is a $40$-dimensional vector.

\begin{figure}[t]
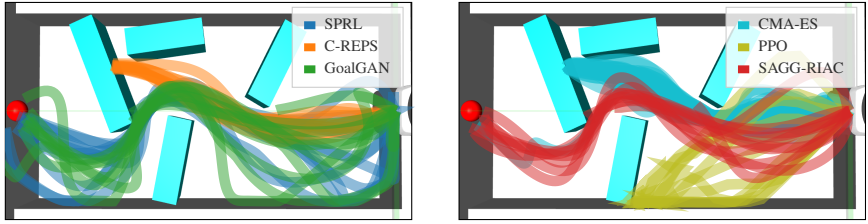

	\centering
	\input{./avoid-final-policies-1.tex}
	\hspace{10pt}
	\input{./avoid-final-policies.tex}
	\caption[Gate Policies]{Trajectories generated by the final policies learned with SPRL, C-REPS, CMA-ES, GoalGAN, SAGG-RIAC and PPO in the reacher environment. The trajectories should reach the red dot while avoiding the cyan boxes. Please note that the visualization is not completely accurate, as we did not account for the viewpoint of the simulation camera when plotting the trajectories.}
	\label{fig:avoid-policies}
	\vspace{-10pt}
\end{figure}

Looking at Figure~\ref{fig:avoid-experiments}, we can see that C-REPS and CMA-ES find a worse optimum compared to SPRL. This local optimum does -- just as in the previous experiment -- not encode optimal behavior, as we can see in Figure~\ref{fig:avoid-policies}. GoalGAN and SAGG-RIAC tend to find the same optimum as SPRL, however with slower convergence and more variance. The sampling distributions visualized in Figure~\ref{fig:avoid-experiments} indicate that SPRL focuses on easier tasks with smaller obstacle sizes first and then moves on to the harder, desired tasks. This also explains the initially lower performance of SPRL on the target task compared to C-REPS and CMA-ES. Figure~\ref{fig:avoid-policies} also shows that PPO \citep{schulman-ppo}, a step-based reinforcement learning algorithm, is not able to solve the task after the same amount of interaction with the environment, emphasizing the complexity of the learning task.

\subsection*{Sparse Ball-in-a-Cup}

\begin{figure}[t]
	\centering
	\begin{tikzpicture}
		\node (O) {\input{./ball-in-a-cup-default-successes.tex}};
		\node (O0) [above right = -106pt and -2pt of O] {\input{./ball-in-a-cup-default-distribution-evolution.tex}};
		\node [above right = -46pt and -5pt of O0] (O1) {\scalebox{0.09}{\includegraphics{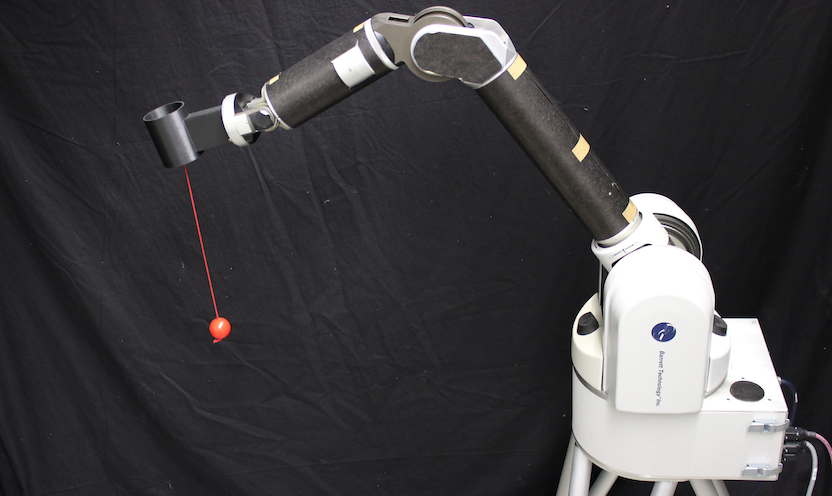}}};
		\node [below = -4pt of O1] (O2) {\scalebox{0.09}{\includegraphics{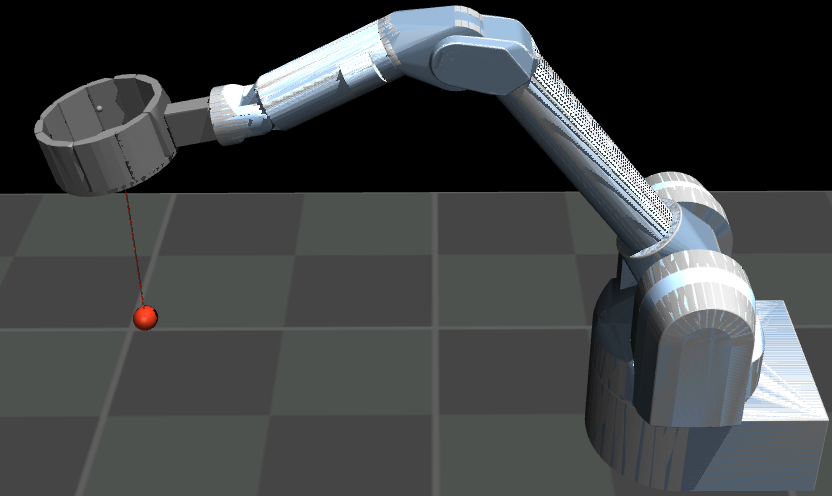}}};
	\end{tikzpicture}
	\vspace{-5pt}
	\caption[BIC Experiments]{The left plots show the $50\%$-quantiles (thick lines) and intervals from $10\%$- to $90\%$-quantile (shaded areas) of the success rate of the investigated algorithms for the sparse Ball-in-a-Cup task. The quantiles were computed from the $10$ best runs out of $20$. One iteration consists of $16$ policy rollouts. Colored areas in the right plot show the sampling distribution $\tilde{\mu}\left(\cvec{c}\right)$ at different iterations of one SPRL run together with the target distribution (black line). The small images on the right visualize the task on the real robot (upper) and in simulation with a scale of $2.5$ (lower).}
	\label{fig:bic-results}
	\vspace{-10pt}
\end{figure}

We conclude the experimental evaluation with a Ball-in-a-Cup task, in which the reward function exhibits a significant amount of sparsity by only returning a reward of $1$ minus an L2 regularization term on the policy parameters, if the ball is in the cup after the policy execution, and $0$ otherwise. The robotic platform is a Barrett WAM, which we simulate using the MuJoCo physics engine \cite{todorov2012mujoco}. The policy again represents a ProMP encoding the desired position of the first, third and fifth joint of the robot. Obviously, achieving the desired task with a poor initial policy is an unlikely event, leading to mostly uninformative rewards and hence a poor learning progress. However, as can be seen in Figure~\ref{fig:bic-results}, giving the learning agent control over the diameter of the cup significantly improves the learning progress by first training with larger cups and only progressively increasing the precision of the movement. Having access to only $16$ samples per iteration, the algorithms did not always learn to achieve the task. However, the final policies learned by SPRL clearly outperform the ones learned by C-REPS, CMA-ES, GoalGAN and SAGG-RIAC. The movements learned in simulation could finally be applied to the robot with a small amount of fine-tuning.

\section{Related Work}

The idea of co-evolving the task together with the learner was explored under different names in various contexts.
In evolutionary robotics, simulation parameters describing a robot were gradually evolved
to match the observations from the real system,
while intermediate controllers were learned entirely in simulation \citep{bongard2004once}.
Recently, this idea got into the spotlight of reinforcement learning
under the name `sim-to-real transfer' \citep{chebotar2018closing}.
In behavioral psychology, a similar concept is known as shaping \citep{skinner1990behavior},
and it has direct links to homotopic-continuation methods \citep{allgower-continuation-methods}
and computational reinforcement learning \citep{erez2008shaping}.
In supervised learning, the paradigm of self-paced curriculum learning
\citep{kumar-self-paced-learning, jiang-self-paced-curriculum-learning}
was successfully applied to automatically determine a sequence of training sets with increasing complexity.
Conceptually closest to ours is the approach
that tackles a much harder problem of learning the \emph{optimal} curriculum \citep{narvekar2018learning},
which turns out to be computationally harder than learning the entire task from scratch at once,
whereas we propose local auxiliary optimization instead as a surrogate for global optimization
that significantly improves sample efficiency of learning.

The process of adaptive task selection can be seen as a form of active learning \citep{settles-active-learning}
since the agent learns how to solve harder tasks by eliciting information from the simpler ones.
Active learning in turn is closely related to curiosity-driven learning \citep{oudeyer2018computational},
which introduced such approaches as
intelligent adaptive curiosity \citep{oudeyer-iac} and intrinsic motivation \citep{barto-intrinsical-motivation}
that suggest focusing learning on tasks that promise high change in reward
based on the recent history of experiences \citep{baranes-sagg-riac}.
Curiosity-driven learning was combined with multi-armed bandit algorithms for automatic task selection in reinforcement learning problems \citep{fabisch-active-contextual-policy-search}
and was applied in robotics to learn goal-reaching movements
with sparse rewards \citep{fournier-accuracy-based-curriculum-learning-rl}.

The idea of reusing knowledge across related tasks
is at the heart of transfer learning in general \citep{pan-transfer-learning-survey}
and transfer in RL in particular \citep{taylor-transfer-learning,lazaric2012transfer}.
Prioritization of tasks for which the agent obtains rewards falling into a certain interval of values
combined with additional reversibility assumptions was shown to enable learning
of high-dimensional object manipulation and maze navigation tasks
\citep{florensa-automatic-goal-generation-for-rl, florensa-reverse-curriculum-for-rl}.
Assuming shared dynamics between tasks and knowledge about the functional form of the reward function
allowed to solve a variety of tasks in the classical computer game Doom \citep{dosovitskiy-learning-to-act}.
Enhanced with universal value function approximators \citep{schaul-uvfas},
reward-based transfer was extremely successful
in robotics applications with sparse reward functions \citep{andrychowicz-her}.

Finally, the investigation of smoothness assumptions on contextual MDPs \citep{modi-cmdps} is highly relevant to our work, as our algorithm implicitly applies such assumptions to restrict the search space by postulating a linear dependency of the policy mean on the context features.

\section{Conclusion}
We proposed a novel approach for generating learning curricula in reinforcement learning problems and developed a practical procedure for simultaneous policy optimization and task sampling distribution adjustment based on an existing information-theoretic contextual policy search algorithm. The progression from `easy' tasks towards the target distribution of `hard' tasks allows to solve problems in which classical contextual policy search algorithms cannot find a satisfying solution.

Although our heuristic of choosing $\alpha$ that controls the trade-off between local improvement and progression towards the desired tasks worked sufficiently well in our experiments, we want to find a more rigorous way of choosing $\alpha$ by e.g. maximizing the learning speed towards the target distribution.

Extensions to step-based policy search algorithms,
such as \citep{peters-reps,schulman-trpo, schulman-ppo},
are conceptually straightforward and are expected to further improve the performance
by leveraging information from every transition in the environment.
Adding a constraint on the lower bound of the policy entropy could furthermore increase the robustness of the algorithm by preserving variance during training on intermediate tasks.

\acknowledgments{Calculations for this research were conducted on the Lichtenberg high performance computer of the TU Darmstadt. This project has received funding from the European Union's Horizon 2020 research and innovation programme under grant agreement No. 640554 (SKILLS4ROBOTS) and from DFG project PA3179/1-1 (ROBOLEAP).}

\bibliography{references}

\clearpage
\appendix
\section{Numerical Stabilization}
We already mentioned that the following optimization problem
\begin{align*}
	\argmax_{p, \tilde{\mu}} \quad & \int_{C, \Theta} R(\cvec{\theta},\cvec{c}) p\left(\cvec{\theta}, \cvec{c}\right) \text{d}\cvec{c} \text{d}\cvec{\theta} - \alpha \text{D}_{\text{KL}}\left(\tilde{\mu} || \mu \right) \\
	\text{s.t.} \quad              & \text{D}_{\text{KL}}\left(p\left(\cvec{\theta}, \cvec{c}\right) || q\left(\cvec{\theta}, \cvec{c}\right)\right) \leq \epsilon                                                         \\
	                               & \int_{C, \Theta} p\left(\cvec{\theta}, \cvec{c}\right) \text{d}\cvec{c} \text{d}\cvec{\theta} = 1                                                                                     \\
	                               & \int_{\Theta} p\left(\cvec{\theta}, \cvec{c}\right) \text{d}\cvec{\theta} = \tilde{\mu}\left(\cvec{c}\right), ~ \forall \cvec{c} \in C
\end{align*}
poses numerical difficulties, especially for small values of $\alpha$. We want to discuss this issue in more detail now by looking at the resulting optimal joint- and context distribution
\begin{align*}
	p^*\left(\cvec{\theta}, \cvec{c}\right) & \propto q\left(\cvec{\theta}, \cvec{c}\right) \exp\left(\frac{A(\cvec{\theta}, \cvec{c})}{\eta}\right)                                      \\
	\tilde{\mu}^*\left(\cvec{c}\right)      & \propto \mu\left(\cvec{c}\right) \exp\left(\frac{V(\cvec{c})}{\alpha}\right)                                                                     \\
	                                        & = \frac{q\left(\cvec{c}\right)}{q\left(\cvec{c}\right)}\mu\left(\cvec{c}\right) \exp\left(\frac{V(\cvec{c})}{\alpha}\right)                      \\
	                                        & = q\left(\cvec{c}\right) \exp\left(\log\left(\frac{\mu\left(\cvec{c}\right)}{q\left(\cvec{c}\right)}\right) + \frac{V(\cvec{c})}{\alpha}\right).
\end{align*}
Note that the expression for $\tilde{\mu}^*$ was re-written to depend on the current sampling distribution $q$ rather than the target distribution $\mu$. This is necessary as the context samples result from sampling $q$.

As already pointed out in the main text, $V(\cvec{c})$ is divided by $\alpha$ in above expression, prohibiting the use of $\alpha$ close to $0$, since the resulting exponential term quickly exceeds the largest number that can be represented using double precision floating point numbers. However, especially these values are crucial for allowing the algorithm to select easy tasks in the first iterations of the algorithm.

Furthermore, the log-term in above expression can also result in large negative numbers if the target distribution $\mu\left(\cvec{c}\right)$ only assigns negligible probability mass to samples from the current sampling distribution $q\left(\cvec{c}\right)$, further amplifying numerical problems.

The dual formulation of the considered optimization problem is
\begin{align*}
	\mathcal{G} & \left(\eta, V\right) = \eta \epsilon + \eta \log\left(\mathbb{E}_q \left[ \exp\left(\frac{A(\cvec{\theta}, \cvec{c})}{\eta}\right)\right]\right) + \frac{\alpha}{\exp\left(1\right)} \mathbb{E}_{q} \left[\exp\left(\log\left(\frac{\mu\left(\cvec{c}\right)}{q\left(\cvec{c}\right)}\right) + \frac{V(\cvec{c})}{\alpha}\right) \right].
\end{align*}
We see that both problematic terms reappear in this dual, making a numerical optimization very hard for small values of $\alpha$ or target distributions that only assign negligible probability mass to the samples of the current sampling distribution.

As already outlined in the main text, adding the two constraints
\begin{align*}
\int_{C} \tilde{\mu}\left(\cvec{c}\right) \text{d}\cvec{c} = 1 \label{eq:mu_norm},\quad \quad \text{D}_{\text{KL}}\left(\tilde{\mu}(\cvec{c}) || q(\cvec{c})\right) \leq \epsilon
\end{align*}
numerically stabilizes the optimization objective by introducing an additional temperature term in the exponential. It can be shown, however, that these additional constraints actually do not further constrain the solution. For this we first note that for the optimal joint- and context distribution $p^*$ and $\tilde{\mu}^*$ it holds that
\begin{align*}
&\int_{C, \Theta} p^*(\cvec{\theta}, \cvec{c}) \mathrm{d}\cvec{\theta} \mathrm{d}\cvec{c} = 1 \Leftrightarrow \int_{C} \left( \int_{\Theta} p^*(\cvec{\theta}, \cvec{c}) \mathrm{d}\cvec{\theta} \right) \mathrm{d}\cvec{c} = 1 \Leftrightarrow \int_{C} \tilde{\mu}^*(\cvec{c}) \mathrm{d}\cvec{c} = 1.
\end{align*}
Furthermore, the following reformulation yields
\begin{align*}
&D_{\mathrm{KL}}(p^*(\cvec{\theta}, \cvec{c}) \| q(\cvec{\theta}, \cvec{c})) \leq \epsilon \\
\Leftrightarrow &\int_{\Theta, C} p^*(\cvec{\theta}, \cvec{c}) \log\left(\frac{p^*(\cvec{\theta}, \cvec{c})}{q(\cvec{\theta}, \cvec{c})}\right) \mathrm{d}\cvec{\theta} \mathrm{d}\cvec{c} \leq \epsilon \\
\Leftrightarrow &\int_{\Theta, C} p^*(\cvec{\theta}, \cvec{c}) \log\left(\frac{\tilde{\mu}^*(\cvec{c})}{q(\cvec{c})}\right) \mathrm{d}\cvec{\theta} \mathrm{d}\cvec{c} + \int_{\Theta, C} p^*(\cvec{\theta}, \cvec{c}) \log\left(\frac{\pi^*(\cvec{\theta} \vert \cvec{c})}{q(\cvec{\theta} \vert \cvec{c})}\right) \mathrm{d}\cvec{\theta} \mathrm{d}\cvec{c} \leq \epsilon \\
\Leftrightarrow &\int_{C} \left(\int_{\Theta} p^*(\cvec{\theta}, \cvec{c}) \mathrm{d}\cvec{\theta} \right) \log\left(\frac{\tilde{\mu}^*(\cvec{c})}{q(\cvec{c})}\right) \mathrm{d}\cvec{c} \leq \epsilon - \int_{\Theta, C} p^*(\cvec{\theta}, \cvec{c}) \log\left(\frac{\pi^*(\cvec{\theta} \vert \cvec{c})}{q(\cvec{\theta} \vert \cvec{c})}\right) \mathrm{d}\cvec{\theta} \mathrm{d}\cvec{c}\\
\Leftrightarrow &\int_{C} \tilde{\mu}^*(\cvec{c}) \log\left(\frac{\tilde{\mu}^*(\cvec{c})}{q(\cvec{c})}\right) \mathrm{d}\cvec{c} \leq \epsilon - E_{\tilde{\mu}^*}\left[\int_{\Theta} \pi^*(\cvec{\theta} \vert \cvec{c}) \log\left(\frac{\pi^*(\cvec{\theta} \vert \cvec{c})}{q(\cvec{\theta} \vert \cvec{c})}\right) \mathrm{d}\cvec{\theta} \right] \\
\Leftrightarrow &D_{\mathrm{KL}}(\tilde{\mu}^*(\cvec{c}) \| q(\cvec{c})) \leq \epsilon - E_{\tilde{\mu}^*} \left[ D_{\mathrm{KL}}(\pi^*(\cvec{\theta} \vert \cvec{c}) \| q(\cvec{\theta} \vert \cvec{c})) \right],
\end{align*}
from which follows that $D_{\mathrm{KL}}(\tilde{\mu}^*(\cvec{c}) \| q(\cvec{c})) \leq \epsilon$, as $E_{\tilde{\mu}^*} \left[ D_{\mathrm{KL}}(\pi^*(\cvec{\theta} \vert \cvec{c}) \| q(\cvec{\theta} \vert \cvec{c})) \right] \geq 0$.

Another aspect of numerical stabilization are the sample weights that are computed after optimizing the dual function. In C-REPS, the weights $w^{\pi}$ are derived from
\begin{align*}
p^*\left(\cvec{\theta}, \cvec{c}\right) & \propto q\left(\cvec{\theta}, \cvec{c}\right) \exp\left(\frac{A(\cvec{\theta}, \cvec{c})}{\eta_p}\right) \\
\Leftrightarrow \pi^*\left(\cvec{\theta} \vert \cvec{c}\right) q\left(\cvec{c}\right) & \propto q\left(\cvec{\theta} \vert \cvec{c}\right) q\left(\cvec{c}\right) \exp\left(\frac{A(\cvec{\theta}, \cvec{c})}{\eta_p}\right) \\
\Leftrightarrow \pi^*\left(\cvec{\theta} \vert \cvec{c}\right) & \propto q\left(\cvec{\theta} \vert \cvec{c}\right) \exp\left(\frac{A(\cvec{\theta}, \cvec{c})}{\eta_p}\right).
\end{align*}
In SPRL, the weights $w^{\pi}$ would slightly differ, as now $p^*\left(\cvec{\theta}, \cvec{c}\right) = \pi^*\left(\cvec{\theta} \vert \cvec{c}\right) \tilde{\mu}^*\left(\cvec{c}\right)$ and hence it holds that
\begin{align*}
\pi^*\left(\cvec{\theta} \vert \cvec{c}\right) & \propto q\left(\cvec{\theta} \vert \cvec{c}\right) \frac{q\left(\cvec{c}\right)}{\tilde{\mu}^*\left(\cvec{c}\right)} \exp\left(\frac{A(\cvec{\theta}, \cvec{c})}{\eta_p}\right) \\
& \propto q\left(\cvec{\theta} \vert \cvec{c}\right) \exp\left(\frac{A(\cvec{\theta}, \cvec{c})}{\eta_p} - \frac{\beta\left(\cvec{c}\right)}{\alpha + \eta_{\mu}}\right).
\end{align*}
However, the second term in the exponential had significantly larger magnitudes in our experiments, in turn leading to poor policy updates. Because of this, we decided to use the same policy update as for the regular C-REPS algorithm. Further investigation of this problem and how to regularize it may allow to use the derived weights instead of the ones from the C-REPS algorithm.

\section{Regularized Policy Updates}

In order to enforce the KL-Bound $\text{D}_{\text{KL}}\left(p(\cvec{\theta}, \cvec{c}) || q(\cvec{\theta}, \cvec{c}) \right) \leq \epsilon$ on the policy and context distribution not only during the computation of the weights but also during the actual inference of the new policy and context distribution, the default weighted linear regression and weighted maximum likelihood objectives need to be regularized.

Given a dataset of $N$ weighted samples 
\begin{align*}
D &= \left\{(w_i^{\cvec{x}}, w_i^{\cvec{y}}, \cvec{x}_i, \cvec{y}_i) \vert i = 1,\ldots, N \right\},
\end{align*}
with $\cvec{x}_i \in \mathbb{R}^{d_{\cvec{x}}}, \cvec{y}_i \in \mathbb{R}^{d_{\cvec{y}}}$, the task of fitting a joint-distribution
\begin{align*}
p(\cvec{x}, \cvec{y}) &= p_{\cvec{y}}(\cvec{y} \vert \cvec{x}) p_{\cvec{x}}(\cvec{x}) = \mathcal{N}(\cvec{y} \vert \cvec{A} \phi(\cvec{x}), \cvec{\Sigma}_{\cvec{y}}) \mathcal{N}(\cvec{x} \vert \cvec{\mu}_{\cvec{x}}, \cvec{\Sigma}_{\cvec{x}})
\end{align*}
to $D$ while limiting the change with regards to a reference distribution
\begin{align*}
q(\cvec{x}, \cvec{y}) &= q_{\cvec{y}}(\cvec{y} \vert \cvec{x}) q_{\cvec{x}}(\cvec{x}) = \mathcal{N}(\cvec{y} \vert \tilde{\cvec{A}} \phi(\cvec{x}), \tilde{\cvec{\Sigma}}_{\cvec{y}}) \mathcal{N}(\cvec{x} \vert \tilde{\cvec{\mu}}_{\cvec{x}}, \tilde{\cvec{\Sigma}}_{\cvec{x}}),
\end{align*}
with feature function $\phi: \mathbb{R}^{d_x} \mapsto \mathbb{R}^{o}$, can be expressed as a constrained optimization problem
\begin{align*}
\max_{\cvec{A}, \cvec{\Sigma}_{\cvec{y}}, \cvec{\mu}_{\cvec{x}}, \cvec{\Sigma}_{\cvec{x}}} &\sum_{i = 1}^N ( w^{\cvec{x}}_i \log(p_{\cvec{x}}(\cvec{x}_i)) + w^{\cvec{y}}_i \log(p_{\cvec{y}}(\cvec{y}_i \vert \cvec{x}_i)) ) \\
\mathrm{s.t.}\ & D_{\mathrm{KL}}(q \| p) \leq \epsilon \\
= \max_{\cvec{A}, \cvec{\Sigma}_{\cvec{y}}, \cvec{\mu}_{\cvec{x}}, \cvec{\Sigma}_{\cvec{x}}} &\sum_{i = 1}^N ( w^{\cvec{x}}_i \log(p_{\cvec{x}}(\cvec{x}_i)) + w^{\cvec{y}}_i \log(p_{\cvec{y}}(\cvec{y}_i \vert \cvec{x}_i)) ) \\
\mathrm{s.t.}\ & E_{q_{\cvec{x}}} \left[ D_{\mathrm{KL}}(q_{\cvec{y}} \| p_{\cvec{y}}) \right] + D_{\mathrm{KL}}(q_{\cvec{x}} \| p_{\cvec{x}}) \leq \epsilon \\
\approx\max_{\cvec{A}, \cvec{\Sigma}_{\cvec{y}}, \cvec{\mu}_{\cvec{x}}, \cvec{\Sigma}_{\cvec{x}}} &\sum_{i = 1}^N \left( w^{\cvec{x}}_i \log(p_{\cvec{x}}(\cvec{x}_i)) + w^{\cvec{y}}_i \log(p_{\cvec{y}}(\cvec{y}_i \vert \cvec{x}_i)) \right) \\
\mathrm{s.t.}\ & \frac{1}{N} \sum_{i = 1}^N D_{\mathrm{KL}}(q_{\cvec{y}}(\cdot \vert \cvec{x}_i) \| p_{\cvec{y}}(\cdot \vert \cvec{x}_i)) + D_{\mathrm{KL}}(q_{\cvec{x}} \| p_{\cvec{x}}) \leq \epsilon.
\end{align*}
Since the distributions $p_{\cvec{x}}$, $p_{\cvec{y}}$, $q_{\cvec{x}}$ and $q_{\cvec{y}}$ are Gaussians, the KL-Divergences can be expressed analytically. Setting the derivative of the Lagrangian with respect to the optimization variables to zero yields to following expressions of the optimization variables in terms of the multiplier $\eta$ and the samples from $D$
\begin{align*}
\cvec{A} &= \left[ \sum_{i = 1}^N \left(w_i \cvec{y}_i + \frac{\eta}{N} \tilde{\cvec{A}} \phi(\cvec{x}_i) \right) \phi(\cvec{x}_i)^T \right] \left[ \sum_{i = 1}^N \left(w_i + \frac{\eta}{N}\right) \phi(\cvec{x}_i) \phi(\cvec{x}_i)^T \right]^{-1}, \\
\\
\cvec{\Sigma}_{\cvec{y}} &= \frac{\sum_{i = 1}^N w_i \Delta\cvec{y}_i \Delta\cvec{y}_i^T + \eta \tilde{\cvec{\Sigma}}_{\cvec{y}}+ \frac{\eta}{N} \Delta\cvec{A} \sum_{i = 1}^N \phi(\cvec{x}_i) \phi(\cvec{x}_i)^T \Delta\cvec{A}^T}{\sum_{i = 1}^N w_i + \eta}, \\
\\
\cvec{\mu}_{\cvec{x}} &= \frac{\sum_{i =1}^N w_i \cvec{x}_i + \eta \tilde{\cvec{\mu}}_{\cvec{x}}}{\sum_{i = 1}^N w_i + \eta}, \\
\\
\cvec{\Sigma}_{\cvec{x}} &= \frac{\sum_{i = 1}^N w_i (\cvec{x}_i - \cvec{\mu}_{\cvec{x}}) (\cvec{x}_i - \cvec{\mu}_{\cvec{x}})^T + \eta \left( \tilde{\cvec{\Sigma}}_{\cvec{x}} + (\cvec{\mu}_{\cvec{x}} - \tilde{\cvec{\mu}}_{\cvec{x}}) (\cvec{\mu}_{\cvec{x}} - \tilde{\cvec{\mu}}_{\cvec{x}})^T \right)}{\sum_{i = 1}^N w_i + \eta},
\end{align*}
with $\Delta\cvec{y}_i = \cvec{y}_i - \cvec{A} \phi(\cvec{x}_i)$ and $\Delta\cvec{A} = \cvec{A} - \tilde{\cvec{A}}$. Above equations yield a simple way of enforcing the KL-Bound on the joint distribution: Since $\eta$ is zero if the constraint on the allowed KL-Divergence is not active, $\cvec{A}$, $\cvec{\Sigma}_{\cvec{y}}$, $\cvec{\mu}_{\cvec{x}}$ and $\cvec{\Sigma}_{\cvec{x}}$ can be first computed with $\eta = 0$ and only if the allowed KL-Divergence is exceeded, $\eta$ needs to be found by searching the root of
$$
f(\eta) = \epsilon - \frac{1}{N} \sum_{i = 1}^N D_{\mathrm{KL}}(q_{\cvec{y}}(\cdot \vert \cvec{x}_i) \| p_{\cvec{y}}(\cdot \vert \cvec{x}_i)) + D_{\mathrm{KL}}(q_{\cvec{x}} \| p_{\cvec{x}}),
$$
where $p_{\cvec{y}}$ and $p_{\cvec{x}}$ are expressed as given by above formulas and hence implicitly depend on $\eta$. As this is a one-dimensional root finding problem, simple algorithms can be used for this task.

\section{Further Experimental Details}

This section is composed of further relevant details on the experiments, which could not be included in the main text due to space limitations. The first part of this section contains general aspects that appeal to more than one experiment as well as tables with important parameters of SPRL, C-REPS, GoalGAN and SAGG-RIAC for the different environments. The remaining parts introduce and discuss details specific to the individual experiments.

\begin{table}[b]
	\caption[Experimental Details]{Important parameters of SPRL and C-REPS in the conducted experiments. The meaning of the symbols correspond to those presented in the  algorithm from the main text and introduced in this appendix.}
	\label{table:exp-det}
	\vskip 0.15in
	\begin{center}
		\begin{small}
			\begin{sc}
				\begin{tabular}{lccccccr}
					\toprule
					& $\epsilon$ & $n_{\text{samples}}$ & Buffer Size & $\zeta$ & $K_{\alpha}$ & $\kappa$ & $\nu$ \\
					\midrule
					Gate ``Global'' & $0.25$ & $100$ & $10$ & $0.002$ & $140$ & $10$ & $10^{-4}$ \\
					Gate ``Precision'' & $0.4$ & $100$ & $10$ & $0.02$ & $140$ & $10$ & $10^{-4}$ \\
					Reacher & $0.5$ & $50$ & $10$ & $0.15$ & $90$ & $20$ & $10^{-1}$ \\
					Ball-in-a-Cup & $0.35$ & $16$ & $5$ & $3.0$ & $15$ & $-$ & $-$ \\
					\bottomrule
				\end{tabular}
			\end{sc}
		\end{small}
	\end{center}
	\vskip -0.1in
\end{table}

\subsection*{General Aspects}

For the gate and the Reacher experiment, the reward function is given by
 \begin{align*}
	R(\cvec{\theta}, \cvec{c}) & = \kappa\ \exp\left(-\norm{\cvec{x}_f\left(\cvec{\theta}\right) - \cvec{x}_g\left(\cvec{c}\right)}_{2}\right) - \nu \sum_{i=0}^N \cvec{u}_i\left(\cvec{\theta}\right)^T \cvec{u}_i\left(\cvec{\theta}\right),
\end{align*}
where $\cvec{x}_f\left(\cvec{\theta}\right)$ is the position of the point-mass or end-effector at the end of the policy execution, $\cvec{x}_g\left(\cvec{c}\right)$ the desired final position, $\cvec{u}_i\left(\cvec{\theta}\right)$ the action applied at time-step $i$ and $\kappa$ and $\nu$ two multipliers that are chosen for each experiment individually. For the visualization of the success rate as well as the computation of the success indicator for the GoalGAN algorithm, the following definition is used: An experiment is considered to be successful, if the distance between final- and desired state is less than a given threshold $\tau$
\begin{align*}
	\text{Success}\left(\cvec{\theta}, \cvec{c}\right) = \begin{cases}
		1,\ \text{if}\  \norm{\cvec{x}_f\left(\cvec{\theta}\right) - \cvec{x}_g\left(\cvec{c}\right)}_{2} < \tau, \\
		0,\ \text{else}.
	\end{cases}
\end{align*}
For the Gate and Reacher environment, the threshold is fixed to $0.05$, while for the Ball-in-a-Cup environment, the threshold depends on the scale of the cup and the goal is set to be the center of the bottom plate of the cup. 

The policies are chosen to be conditional Gaussian distributions $\mathcal{N}(\cvec{\theta} \vert \cvec{A} \phi(\cvec{c}), \cvec{\Sigma}_{\cvec{\theta}})$, where $\phi(\cvec{c})$ is a feature function. SPRL and C-REPS both use linear policy features in all environments.

In the Reacher and the Ball-in-a-Cup environment, the parameters $\cvec{\theta}$ encode a feed-forward policy by weighting several Gaussian basis functions over time
\begin{align*}
\cvec{u}_i\left(\cvec{\theta}\right) = \cvec{\theta}^T \cvec{\psi}\left(t_i\right), \quad \cvec{\psi}_j\left(t_i\right) = \frac{b_j\left(t_i\right)}{\sum_{l=1}^L b_l\left(t_i\right)}, \quad b_j\left(t_i\right) = \exp\left(\frac{\left(t_i - c_j\right)^2}{2L}\right),
\end{align*}
where the centers $c_j$ and length $L$ of the basis functions are chosen individually for the experiments. With that, the policy represents a Probabilistic Movement Primitive \citep{paraschos-promps}, whose mean and co-variance matrix are progressively shaped by the learning algorithm to encode movements with high reward.

In order to increase the robustness of SPRL and C-REPS while reducing the sample complexity, an experience buffer storing samples of recent iterations is used. The ``size'' of this buffer dictates the number of past iterations, whose samples are kept. Hence, in every iteration, C-REPS and SPRL work with $N_{\text{SAMPLES}} \times \text{BUFFER SIZE}$ samples, from which only $N_{\text{SAMPLES}}$ are generated by the policy of the current iteration.

We use the CMA-ES implementation given by \citep{hansen2019pycma}. As it only allows to specify one initial variance for all dimensions of the search distribution, this variance is set to the maximum of the variances contained in the initial co-variance matrices used by SPRL and C-REPS.

For the GoalGAN algorithm, the percentage of samples that are drawn from the buffer containing already solved tasks is fixed to $20\%$. Since GoalGAN generates the learning curriculum using ``Goals of Intermediate Difficulty'', it is necessary to execute the policy at least twice in each context. Hence, for $30\%$ of the contexts (after subtracting the previously mentioned $20\%$ of ''old`` samples), the policy is executed twice. For $10\%$ of the contexts, the policy is executed four times. The noise added to the samples of the GAN $\epsilon_{\text{NOISE}}$ and the number of iterations that pass between the training of the GAN $\Delta_{\text{TRAIN}}$ are chosen individually for the experiments.

The SAGG-RIAC algorithm requires, besides the probabilities for the sampling modes which are kept as in the original paper, two hyperparameters to be chosen: The maximum number of samples to keep in each region $n_{\text{GOALS}}$ as well as the maximum number of recent samples for the competence computation $n_{\text{HIST}}$. 

Tables \ref{table:exp-det} and \ref{table:exp-det-2} show the aforementioned hyperparameters for the different algorithms as well as reward function parameters for the different environments.

\begin{figure}
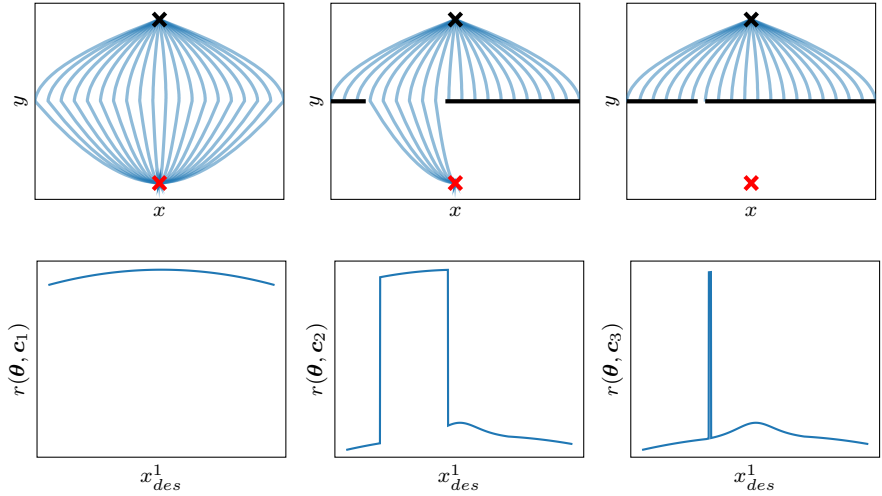

\centering
\begin{tikzpicture}
\node (leftnode) at (0, 0) {\input{gate-examples-20.tex}};
\node [below left = 0.25 and -3.96 of leftnode] {\input{gate-cost-function-20-3.tex}};

\node[right = 0.0 of leftnode] (middlenode) {\input{gate-examples-3.tex}};
\node [below = 0.25 of middlenode] {\input{gate-cost-function-3-3.tex}};

\node[right = 0.0 of middlenode] (rightnode) {\input{gate-examples-0.1.tex}};
\node [below = 0.25 of rightnode] {\input{gate-cost-function-0.1-3.tex}};
\end{tikzpicture}
\caption[Test]{The columns show visualizations of the point-mass trajectories (upper plots) as well as the obtained rewards (lower plots) in the gate task, when the desired position of the first PD-controller is varied while all other parameters are kept fixed such that a stable control law is obtained. In every column, the gate is positioned at $x=4.0$ while the size of it varies from $20$ (left), over $3$ (middle) to $0.1$ (right).}
\label{fig:gate-reward-vis}
\end{figure}

\begin{table}[b]
	\caption[Experimental Details]{Important parameters of GoalGAN and SAGG-RIAC in the conducted experiments. The meaning of the symbols correspond to those introduced in this appendix.}
	\label{table:exp-det-2}
	\vskip 0.15in
	\begin{center}
		\begin{small}
			\begin{sc}
				\begin{tabular}{lccccccr}
					\toprule
					& $\epsilon_{\text{NOISE}}$ & $\Delta_{\text{TRAIN}}$ & $n_{\text{GOALS}}$ & $n_{\text{HIST}}$ \\
					\midrule
					Gate ``Global'' & $0.05$ & $5$ & $100$ & $500$ \\
					Gate ``Precision'' & $0.05$ & $5$ & $100$ & $200$ \\
					Reacher & $0.1$ & $5$ & $80$ & $300$ \\
					Ball-in-a-Cup & $0.05$ & $3$ & $50$ & $120$  \\
					\bottomrule
				\end{tabular}
			\end{sc}
		\end{small}
	\end{center}
	\vskip -0.1in
\end{table}

\subsection*{Gate Experiment}

The linear system that describes the behavior of the point-mass is given by
\begin{equation*}
	\begin{bmatrix}
		\dot{x} \\
		\dot{y}
	\end{bmatrix} = \begin{bmatrix}
		5 \\
		-1
	\end{bmatrix} + \cvec{u} + \cvec{\delta},\quad \cvec{\delta} \sim \mathcal{N}\left(\cvec{0}, 2.5 \times 10^{-3} \cvec{I} \right).
\end{equation*}
The point-mass is controlled by two PD-controllers
\begin{align*}
	\text{PD}_i\left(x, y\right) = \cvec{K}_i \begin{bmatrix}
		x_i - x \\
		y_i - y
	\end{bmatrix} + \cvec{k}_i,\ i\in \left[1, 2\right], \quad \cvec{K}_1, \cvec{K}_2 \in \mathbb{R}^{2 \times 2},\ \cvec{k}_1, \cvec{k}_2 \in \mathbb{R}^2,\ x_1, x_2, y_1, y_2 \in \mathbb{R},
\end{align*}
where $x$ is the $x$-position of the point mass and $y$ its position on the $y$-axis. In initial iterations of the algorithm, the sampled PD-controller parameters sometimes make the control law unstable, leading to very large penalties from the L2-regularization of the applied actions and hence to numerical instabilities in SPRL and C-REPS because of very large negative rewards. Because of this, the reward is clipped to always be above $0$.

Table~\ref{table:exp-det} shows that a large number of samples per iteration for both the ``global'' and ``precision'' setting are used. This is purposefully done to keep the influence of the sample size on the algorithm performance as low as possible, as both of these settings serve as a first conceptual benchmark of our algorithm.

Figure~\ref{fig:gate-reward-vis} helps in understanding, why SPRL drastically improves upon C-REPS especially in the ``precision'' setting, even with this large amount of samples. For narrow gates, the reward function has a local maximum which tends to attract both C-REPS and CMA-ES, as the chance of sampling a reward close to the true maximum is very unlikely. By first training on contexts in which the global maximum is more likely to be observed and only gradually moving towards the desired contexts, SPRL avoids this sub-optimal solution.

\subsection*{Reacher Experiment}

In the Reacher experiment, the ProMP encoded by the policy $\pi$ has $20$ basis functions of width $L=0.03$. The centers are evenly spread in the interval $[-0.2, 1.2]$ and the time interval of the movement is normalized to lie in the interval $[0, 1]$ when computing the activations of the basis functions. Since the robot can only move within the $xy$-plane, $\cvec{\theta}$ is a $40$-dimensional vector. As we can see in Table~\ref{table:exp-det}, the number of samples in each iteration was decreased to $50$, which in combination with the increased dimensionality of $\cvec{\theta}$ makes the task more challenging. As a side-effect, this reduced the training time, as less simulations need to be executed during training.

While working on the Reacher experiment, we realized that ensuring a minimum amount of variance of the intermediate context distributions of SPRL stabilizes the learning, as for very narrow context distributions, the progression of the algorithm towards the target context distribution becomes very slow.  As soon as the current context distribution is sufficiently close to the target one, which in the Reacher experiment is considered to be the case if the KL-Divergence between intermediate- and target context distribution drops below a value of $20$, the context distribution is allowed to contract without restrictions. Before that happens, the variance is each dimension is lower-bounded by $3 \times 10^{-5}$.

The PPO results are obtained using the version from \citep{stable-baselines}, for which a step-based version of the Reacher experiment is used, in which the reward function is given by
$$
r(\cvec{s}, \cvec{a}) = \exp \left(-2.5 \sqrt{(x - x_g)^2 + (y - y_g)^2}\right),
$$
where $\cvec{s} = (x\ \dot{x}\ y\ \dot{y})$ is the position and velocity of the end-effector, $\cvec{a} = (a_x\ a_y)$ the desired displacement of the end-effector (just as in the regular Reacher task from the OpenAI Gym simulation environment) and $x_g$ and $y_g$ is the $x-$ and $y-$ position of the goal. When an obstacle is touched, the agent is reset to the initial position. This setup led to the best performance of PPO, while resembling the structure of the episodic learning task used by the other algorithms (a version in which the episode ends as soon as an obstacle is touched led to a lower performance of PPO).

To ensure that the poor performance of PPO is not caused by an inadequate choice of hyperparameters, PPO was run on an easy version of the task in which the two obstacle sizes were set to $0.01$, where it encountered no problems in solving the task.

Every iteration of PPO uses $3600$ environment steps, which corresponds to $24$ trajectory executions in the episodic setting. PPO uses an entropy coefficient of $10^{-3}$, $\gamma=0.999$ and $\lambda=1$. The neural network that learns the value function as well as the policy has two dense hidden layers with $164$ neurons and $\tanh$ activation functions. The number of minibatches is set to $5$ while the number of optimization epochs is set to $15$. The standard deviation in each action dimension is initialized to $1$, giving the algorithm enough initial variance, as the actions are clipped to the interval $[-1, 1]$ before being applied to the robot.

\subsection*{Ball-in-a-Cup Experiment}

For the Ball-in-a-Cup environment, the $9$ basis functions  of the ProMP are spread over the interval $[-0.01, 1.01]$ and have width $L=0.0035$. Again, the time interval of the movement is normalized to lie in the interval $[0,1]$ when computing the basis function activations. The ProMP encodes the offset of the desired position from the initial position.
By setting the first and last two basis functions to $0$ in each of the three dimensions, the movement always starts in the initial position and returns to it after the movement execution. All in all, $\theta$ is a $15$-dimensional vector.

The reward function is defined as
\begin{align*}
R(\cvec{\theta}, \cvec{c}) & = \begin{cases}%
1 - 0.07 \cvec{\theta}^T \cvec{\theta}&,\ \text{if successful} \\
0&,\ \text{else}%
\end{cases},
\end{align*}
encoding a preference over movements that deviate as less as possible from the initial position while still solving the task.

Looking back at Table~\ref{table:exp-det}, the value of $\zeta$ stands out, as it is significantly higher than in the other experiments. We suppose that such a large value of $\zeta$ is needed because of the shape of the reward function, which creates a large drop in reward if the policy is sub-optimal. Because of this, the incentive required to encourage the algorithm to shift probability mass towards contexts in which the current policy is sub-optimal needs to be significantly higher than in the other experiments.

Just as for the Reacher experiment, we also lower-bound the variance of the intermediate context distributions to $0.01$ until the KL-Divergence between intermediate and target distribution falls below a value of $200$.

After learning the movements in simulation, the successful runs were executed on the real robot. Due to simulation bias, just replaying the trajectories did not work satisfyingly. At this stage, we could have increased the variance of the movement primitive and re-trained on the real robot. As sim-to-real transfer is, however, not the focus of this paper, we decided to manually adjust the execution speed of the movement primitive by a few percent, which yielded the desired result.

\end{document}